\documentclass[acmtog,nonacm]{acmart}

\usepackage{booktabs} 
\usepackage{hyperref}
\usepackage{cleveref}

\usepackage[ruled]{algorithm2e} 

\SetAlFnt{\small}
\SetAlCapFnt{\small}
\SetAlCapNameFnt{\small}
\SetAlCapHSkip{0pt}

\begin{document}
\title{Style-NeRF2NeRF: 3D Style Transfer from Style-Aligned Multi-View Images}

\author{Haruo Fujiwara}
\orcid{0009-0001-0443-4477}
\affiliation{%
  \institution{The University of Tokyo}
  \country{Japan}}
\email{fujiwara@mi.t.u-tokyo.ac.jp}

\author{Yusuke Mukuta}
\orcid{0000-0002-7727-5681}
\affiliation{%
  \institution{The University of Tokyo}
  \country{Japan}}
\affiliation{%
  \institution{RIKEN}
  \country{Japan}}
\email{mukuta@mi.t.u-tokyo.ac.jp}

\author{Tatsuya Harada}
\orcid{0000-0002-3712-3691}
\affiliation{%
  \institution{The University of Tokyo}
  \country{Japan}}
\affiliation{%
  \institution{RIKEN}
  \country{Japan}}
\email{harada@mi.t.u-tokyo.ac.jp}



\begin{abstract}
We propose a simple yet effective pipeline for stylizing a 3D scene, harnessing the power of 2D image diffusion models. Given a NeRF model reconstructed from a set of multi-view images, we perform 3D style transfer by refining the source NeRF model using stylized images generated by a style-aligned image-to-image diffusion model.
Given a target style prompt, we first generate perceptually similar multi-view images by leveraging a depth-conditioned diffusion model with an attention-sharing mechanism. Next, based on the stylized multi-view images, we propose to guide the style transfer process with the sliced Wasserstein loss based on the feature maps extracted from a pre-trained CNN model.
Our pipeline consists of decoupled steps, allowing users to test various prompt ideas and preview the stylized 3D result before proceeding to the NeRF fine-tuning stage.
We demonstrate that our method can transfer diverse artistic styles to real-world 3D scenes with competitive quality. Result videos are also available on our project page: \url{https://haruolabs.github.io/style-n2n/}
\end{abstract}

%
%

\ccsdesc[500]{Computing methodologies~Non-photorealistic rendering}
\ccsdesc[500]{Computing methodologies~Computer vision representations}

%
%

\keywords{Neural Radiance Fields, Neural Rendering, Style Transfer, Diffusion Model}

\begin{teaserfigure}
  \includegraphics[width=\textwidth]{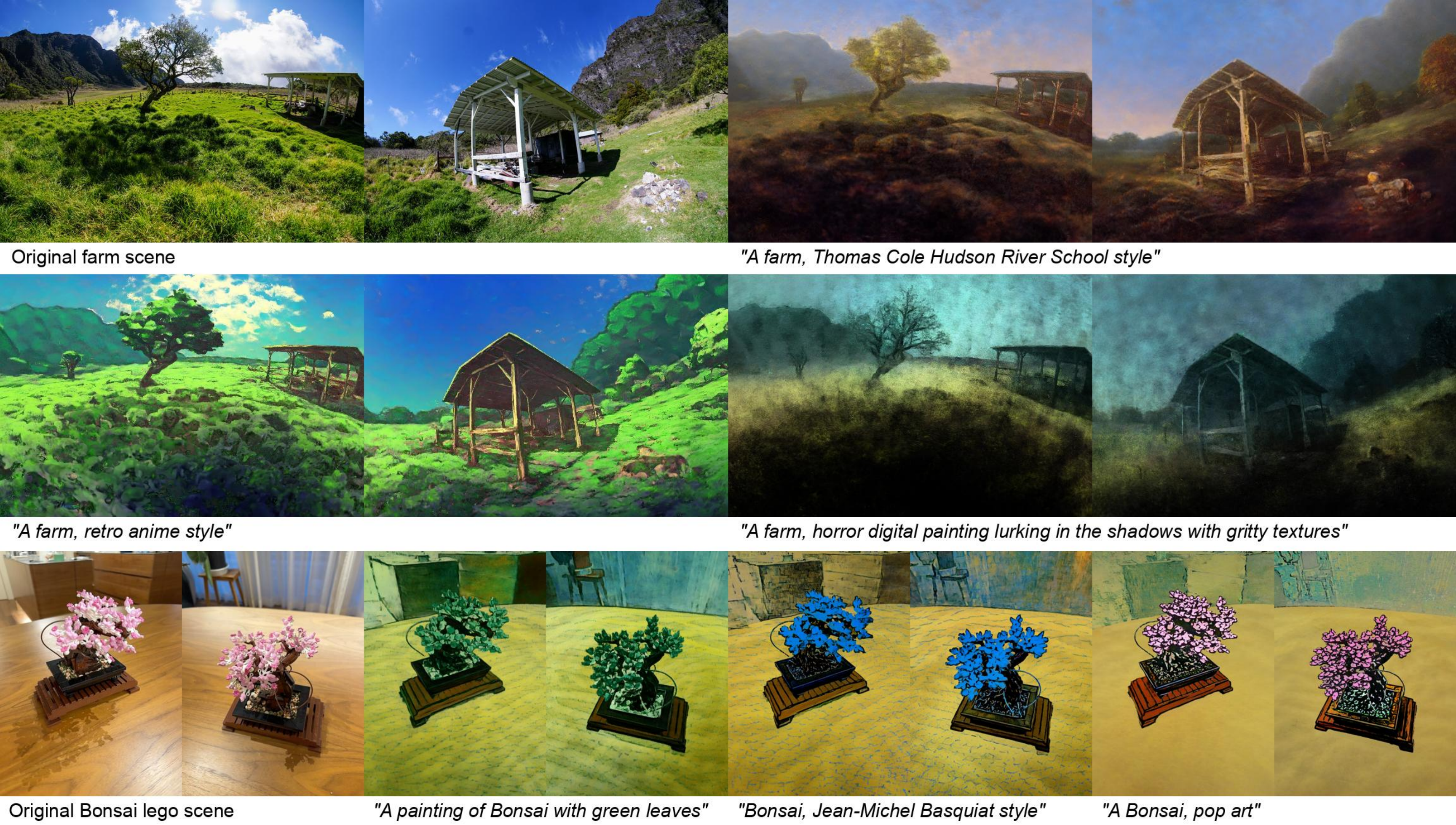}
  \caption{Our method makes it possible to perform 3D artistic style transfer on a pre-trained NeRF scene using text descriptions.}
  \label{fig:teaser}
\end{teaserfigure}

\maketitle

\section{Introduction}
\label{sec:intro}
Thanks to recent advancements in 3D reconstruction techniques such as Neural Radiance Fields (NeRF) \cite{mildenhall2020nerf}, it is nowadays possible for creators to develop a 3D asset or a scene from captured real-world data without intensive labor.
While such 3D reconstruction methods work well, editing an entire 3D scene to match a desired style or concept is not straightforward.

For instance, editing conventional 3D scenes based on explicit representations like mesh often involves specialized tools and skills. Changing the appearance of the entire mesh-based scene would often require skilled labor, such as shape modeling, texture creation, and material parameter modifications.

At the advent of implicit 3D representation techniques such as NeRF, style editing methods for 3D are also emerging \cite{nguyen2022snerf,wang2023nerfart,liu2023stylerf, kamata2023instruct,instructnerf2023,dong2024vica} to enhance creators’ content development process.
Following the recent development of 2D image generation models, prominent works such as Instruct-NeRF2NeRF \cite{instructnerf2023,igs2gs} and ViCA-NeRF \cite{dong2024vica} proposed to leverage the knowledge of large-scale pre-trained text-to-image (T2I) models to supervise the 3D NeRF editing process.

These methods employ a custom pipeline based on an instruction-based T2I model ”Instruct-Pix2Pix” \cite{brooks2023instructpix2pix} to stylize a 3D scene with text instructions. While Instruct-NeRF2NeRF is proven to work well for editing 3D scenes including large-scale 360 environments, their method involves an iterative process of editing and replacing the training data during NeRF optimization, occasionally resulting in unpredictable results. As editing by Instruct-Pix2Pix runs in tandem with NeRF training, we found adjusting or testing editing styles beforehand difficult.

To overcome this problem, we propose an artistic style-transfer method that trains a source 3D NeRF scene on stylized images \emph{prepared in advance} by a text-guided style-aligned diffusion model. Training is guided by \emph{Sliced Wasserstein Distance} (SWD) loss \cite{heitz2021sliced,li2022sliced} to effectively perform 3D style transfer with NeRF.
A summary of our contributions is as the follows:

\begin{itemize}
\item We propose a novel 3D style-transfer approach for NeRF, including large-scale outdoor scenes.
\item We show that a style-aligned diffusion model conditioned on depth maps of corresponding source views can generate \emph{perceptually} view-consistent style images for fine-tuning the source NeRF. Users can test stylization ideas with the diffusion pipeline before proceeding to the NeRF fine-tuning phase.
\item We find that fine-tuning the source NeRF with SWD loss can perform 3D style transfer well.
\item Our experimental results illustrate the rich capability of stylizing scenes with various text prompts.
\end{itemize}

\section{Related Work}
\subsection{Implicit 3D Representation}
NeRF, introduced by the seminal paper \cite{mildenhall2020nerf}, became one of the most popular implicit 3D representation techniques due to several benefits. NeRF can render photo-realistic novel views with arbitrary resolution due to its continuous representation with a compact model compared to explicit representations such as polygon mesh or voxels. In our research, we use the "nerfacto" model implemented by Nerfstudio \cite{tancik2023nerfstudio}, which is a combination of modular features from multiple papers \cite{wang2021nerfminus,barron2022mip,muller2022instant,martin2021nerfw,verbin2022ref}
, designed to achieve a balance between speed and quality.

\subsection{Diffusion Models}
Diffusion models \cite{sohl2015deep,song2020score,dhariwal2021diffusion} are generative models that have gained significant attention for their ability to generate high-quality, diverse images. Inspired by classical non-equilibrium thermodynamics, they are trained to generate an image by reversing the diffusion process, progressively denoising noisy images towards meaningful ones. Diffusion models are commonly trained with classifier-free guidance \cite{ho2022classifier} to enable image generation conditioned on an input text.

\subsubsection{Controlled Generations with Diffusion Models.}
Leveraging the success of T2I diffusion models, recent research has expanded their application to controlled image generation and editing, notably in image-to-image (I2I) tasks \cite{meng2021sdedit,parmar2023zero_i2i,kawar2023imagic,tumanyan2023plug,mokady2023null,hertz2023delta,hertz2022prompt, brooks2023instructpix2pix}. For example, SDEdit \cite{meng2021sdedit} achieves this by first adding noise to a source image and then guiding the diffusion process toward an output based on a given prompt. ControlNet \cite{zhang2023adding} was proposed as an add-on architecture for training T2I diffusion models with extra conditioning inputs such as depth, pose, edge maps, and more. Several recent techniques \cite{hertz2023style, sohn2024styledrop,cheng2023general_i2i} focus on generating style-aligned images.
In our work, we use a depth-conditioned I2I pipeline with an attention-sharing mechanism similar to "StyleAligned" \cite{hertz2023style} to create a set of multi-view images sharing a consistent style.

\subsection{Style Transfer}
\subsubsection{2D Style Transfer.}
Style transfer originally refers to a technique for blending images, a source image and a style image, to create another image that retains the first's content but exhibits the second's style.
Since the introduction of the foundational style transfer algorithm proposed by \cite{gatys2015neural}, many follow-up works for 2D style transfer have been explored for further improvements such as faster optimization \cite{johnson2016perceptual_faststyle,huang2017arbitrary_adain}, zero-shot style-transfer \cite{li2017universal}, and photo-realism \cite{luan2017deepphoto}. 
Furthermore, content stylization methods using only text descriptions for style \cite{frenkel2024implicit,sohn2024styledrop, shah2023ziplora} are showing promising results due to the recent progress in controllable diffusion models.

\subsubsection{3D Style Transfer.}
Several recent 3D style transfer works have applied style transfer techniques using deep feature statistics to NeRF \cite{liu2023stylerf, wang2023nerfart,zhang2022arf,chiang2022stylizing,huang2022stylizednerf,nguyen2022snerf,pang2023locally}. 
In addition to such stylization methods based on a style reference, text-driven 3D editing techniques leveraging foundational 2D Text-to-Image (T2I) models are developed.
While Instruct 3D-to-3D \cite{kamata2023instruct} proposed using Score Distillation Sampling (SDS) loss \cite{poole2022dreamfusion} for text guided NeRF stylization, Instruct-NeRF2NeRF \cite{instructnerf2023} and ViCA-NeRF \cite{dong2024vica} perform NeRF editing by optimizing the underlying scene with a process referred to as Iterative Dataset Update (Iterative DU), which gradually replaces the input images with edited images from InstructPix2Pix \cite{instructnerf2023}, an image-conditioned instruction-based diffusion model, followed by an update of NeRF. Inspired by these methods, we also develop a 3D style transfer method for NeRF, supervised by images created by a diffusion pipeline but \textit{without} Iterative DU.


\begin{figure*}[h]
  \centering
  \includegraphics[width=\textwidth]{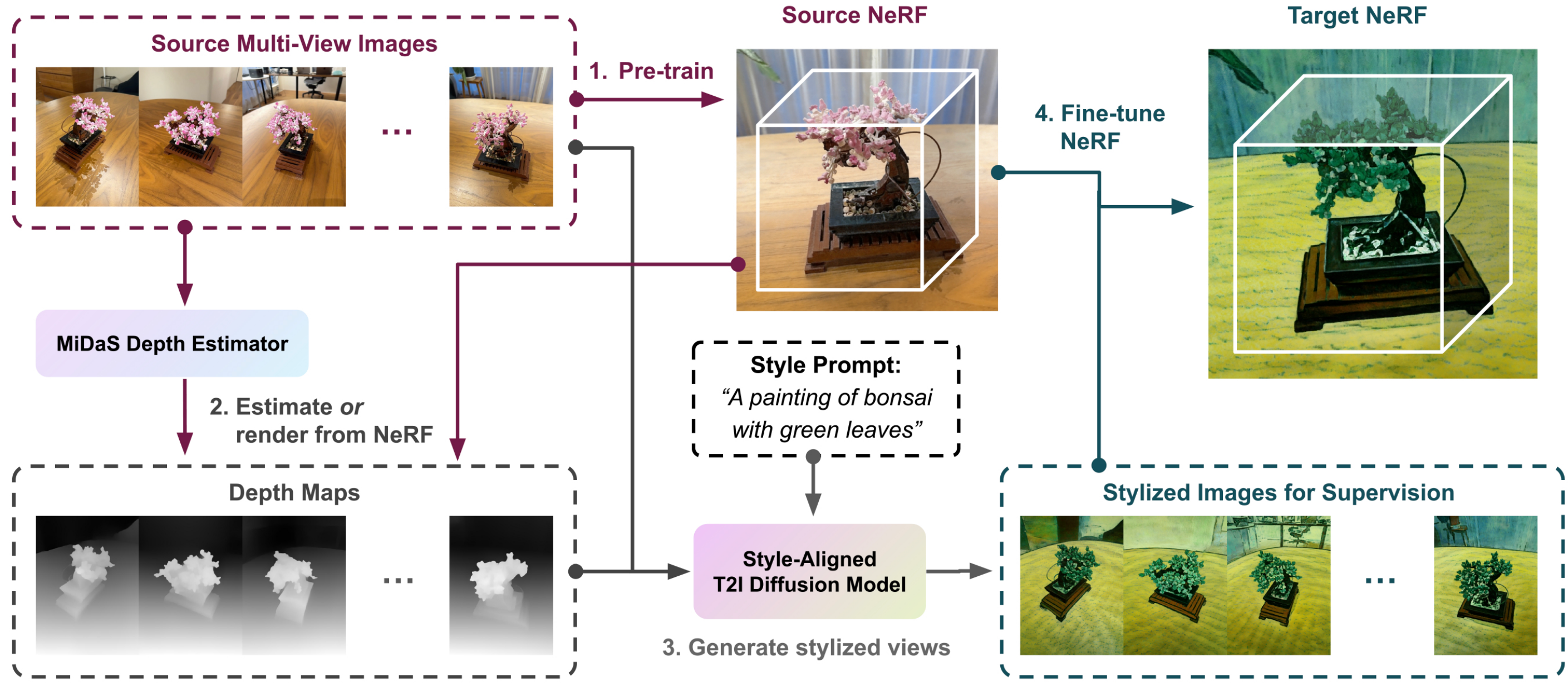}
  \caption{ \textbf{Overall Pipeline.} Our method consists of distinct procedures. We first prepare a NeRF model of the source view images. Given the depth maps of the corresponding views (by either estimation or rendering by NeRF), we generate stylized multi-view images using a style-aligned diffusion model. Lastly, we fine-tune the source NeRF on the stylized images using the SWD loss.
  }
  \label{fig:pipeline}
\end{figure*}

\section{Method}
\subsection{Preliminaries}
\subsubsection{Neural Radiance Fields.}
NeRF \cite{mildenhall2020nerf} models a volumetric 3D scene as a continuous function by mapping a 3D coordinate $\mathbf{x}= (x, y, z)$ and a 2D viewing direction $\mathbf{d} = (\theta, \phi)$ to a color (RGB) $\mathbf{c}$ and a density ($\sigma$). This function $F_\theta: (\mathbf{x}, \mathbf{d}) \mapsto (\mathbf{c},\mathbf{\sigma}) $ is often parameterized by a neural network, a voxel grid structure \cite{fridovich2022plenoxels}, or a hybrid representation to accelerate performance \cite{muller2022instant, sun2022direct, sun2022improved}. Given a NeRF model trained on a set of 2D images taken from various viewpoints of a target scene, the accumulated color $C(\mathbf{r})$ along an arbitrary camera ray $\mathbf{r}(t) = \mathbf{o} + t\mathbf{d}$ is calculated with the quadrature rule by volume rendering \cite{max1995optical}:

\begin{equation}
C(\mathbf{r}) = \sum_{k=1}^{K} T_k \bigl(1 - \exp({-\sigma_k \delta_k})\bigr) \mathbf{c}_k,
T_k = \exp{ \Bigl( -\sum_{j=1}^{k-1} \sigma_j \delta_j \Bigr) }
\label{eq:vol-render}
\end{equation}
where $\delta_k = t_{k+1} - t_k$ is the distance between sampled points on the ray and $T_k$ is the accumulated transmittance from origin $\mathbf{o}$ to the $k$-th sample.

\subsubsection{Conditional Diffusion Models.}
Recent T2I diffusion models \cite{rombach2022high,podell2023sdxl} are built with a U-net architecture \cite{ronneberger2015u} integrated with convolutional layers and attention blocks \cite{vaswani2017attention}. Within the model, attention blocks play a crucial role in correlating text with relevant parts of the deep features during image generation. Our work uses an open-source latent diffusion model \cite{podell2023sdxl}, which includes a CLIP text encoder \cite{radford2021learning} for text embedding. The cross-attention between contextual text embedding and the deep features of the denoising network is calculated as follows:

\begin{equation}
\texttt{Attn}(Q, K, V) = \texttt{softmax} \left( \frac{QK^T}{\sqrt{d_k}}V \right)
\label{eq:eq-attention}
\end{equation}
where $Q \in m \times d_k, K \in m \times d_k, V \in m \times d_h$ are projection matrices for a deep feature map $\phi \in \mathbb{R}^{m \times d_h}$. We may interpret the attention operation in equation \ref{eq:eq-attention} as values $V$,  originating from conditional text, weighted by the correlation of queries $Q$, and the keys $K$. There are often $N_h$ multiple attention heads in each layer along the $d_h$ dimension to allow the model to attend to information from different subspaces in feature space jointly:

\begin{equation}
\texttt{MultiHeadAttn}(Q, K, V) = \underset{j \in N_h}{\texttt{concat}} 
\Bigl[ \texttt{Attn} (Q^{(j)}, K^{(j)}, V^{(j)}) \Bigr]
\label{eq:eq-multi-head-attention}
\end{equation}

\subsection{Style-NeRF2NeRF}
Our method is a distinct two-step process. First, we prepare stylized images of corresponding source views using our style-aligned diffusion pipeline, and then refine the source NeRF model based on the generated views to acquire a style-transferred 3D scene.

\label{style-aligned i2i}
\subsubsection{Style-Aligned Image-to-Image Generation.} \label{style-aligned i2i}
Given a set of source view images $\{I^{(i)}\} (i=1,\ldots,N)$, our first goal is to generate a corresponding set of stylized view images $I^{(i)}_{c} = U_{\theta }(I^{(i)}, c)$ under a text condition $c$ with as much perceptual view consistencies among images where $U_\theta$ consists of a sampling process such as DDIM \cite{song2020denoising}.

Although T2I diffusion models can generate rich images with arbitrary text prompts, merely sharing the same prompt across different source views is insufficient to generate stylized images with a perceptually consistent style.
To alleviate this problem, we apply a fully-shared-attention variant of a style-aligned image generation method proposed by \cite{hertz2023style}. Let  $Q_i, K_i, V_i$ be the queries, keys, and values from a deep feature $\phi_i$ for view $I^{(i)}$, then we generate $n$ stylized views simultaneously using the following fully-shared-attention:

\begin{equation}
\texttt{Attn}(Q_i, K_{1\ldots n}, V_{1\ldots n}) \\
\label{eq:important}
\end{equation}
\begin{equation}
K_{1\ldots n} = [K_1, K_2, \ldots K_n]^T, V_{1\ldots n} = [V_1, V_2, \ldots V_n]^T
\label{eq:eq-shared-attention}
\end{equation}

Figure \ref{fig:style-align-vs-no-style-align} illustrates an example of multi-view images generated with and without the fully-shared-attention mechanism.

\subsubsection{Conditioning on Source Views.}
To further strengthen perceptual consistencies across multi-view frames, we attach a depth-conditioned ControlNet \cite{zhang2023adding} and optionally enable SDEdit \cite{meng2021sdedit} for conditioning on the source view. 
As for the depth inputs, we may either render the corresponding depth maps from the source NeRF or use an off-the-shelf depth estimator model such as MiDaS \cite{ranftl2020towards}.

Given a set of translated multi-view images based on style text and their corresponding camera poses for training a source NeRF model, we may proceed to the NeRF refining stage described below.

\subsubsection{NeRF Fine-Tuning.} \label{training}
Based on the \textit{perceptually} view-consistent images $\{I^{(i)}_c\}$ created by the style-aligned image-to-image diffusion model, our next objective is to fine-tune the source NeRF scene to reflect the target style in a 3D consistent manner.

Although the stylized multi-view images are a good starting point for fine-tuning the source NeRF, we found that using a common RGB pixel loss is prone to over-fitting due to ambiguities in 3D geometry and color. Therefore, an alternative loss function that reflects the \textit{perceptual} similarity is preferred for guiding the 3D style-transfer process. To meet our requirement, we employ the \emph{Sliced Wasserstein Distance loss} (SWD loss) \cite{heitz2021sliced}.

\subsection{Sliced Wasserstein Distance Loss.}
Feature statistics of pre-trained Convolutional Neural Networks (CNNs) such as VGG-19 \cite{simonyan2014very} are known to be useful for representing a style of an image \cite{gatys2015neural,johnson2016perceptual_faststyle,huang2017arbitrary_adain,li2017universal,luan2017deepphoto}. In our study we employ the SWD loss originally proposed for texture synthesis \cite{heitz2021sliced} as the loss term to guide the style-transfer process for NeRF.

Let $F^l_m \in \mathbb{R}^{N_l} (m=1,\ldots,M_l)$ denote the feature vector of the $l$-th convolutional layer at pixel $m$ where $M_l$ is the number of pixels and $N_l$ is the feature dimension size. Using the delta Dirac function, we may express the discrete probability density function $p^l(x)$ of the features for layer $l$ as below:

\begin{equation}
    p^l(x) = \frac{1}{M_l} \sum_{m=1}^{M_l} \delta_{F^l_m}(x)
\label{eq:feature-stats}
\end{equation}

Using the feature distributions $p^l, \hat{p}^l$ for image $I$ and its corresponding optimization target $\hat{I}$, the style loss is defined as a sum of SWD over the layers:

\begin{equation}
    \mathcal{L}_{style} = \sum_{l=1}^{L} \mathcal{L}_{SWD}(p^l, \hat{p}^l)
\end{equation}
where, $\mathcal{L}_{SWD}$ is the SWD term defined as the expectation over $1$-dimensional Wasserstein distances of features projected by random directions $V \in \mathcal{S}^{N_l-1}$ sampled from a unit hypersphere. 

\begin{figure}[h]
  \includegraphics[width=\columnwidth]{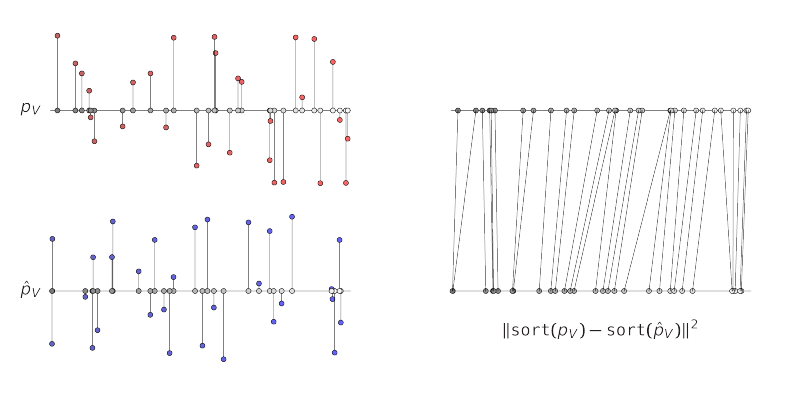}
  \caption{ \textbf{Sliced Wasserstein Distance.} $p$ and $\hat{p}$ are projected onto a random unit direction $V$ (left). The $1$-dimensional Wasserstein distance can be calculated by taking the $L^2$ difference between the sorted projections $p$ and $\hat{p}$ (right). Expectation over random $V$ vectors is a practical approximation of the $N$-dimensional Wasserstein distance.
  }
  \label{fig:sliced-wasserstein}
\end{figure}

Using the projected scalar features $p^l_V = \{ \langle F^l_m, V \rangle \}, \forall m$, where $\langle, \rangle$ denotes a dot product, one may obtain $\mathcal{L}_{SWD}$ as the following where the $1$-dimentional $2$-Wasserstein distance $\mathcal{L}_{SW1D}$ is trivially calculated in a closed form by taking the element-wise $L^2$ distances between sorted scalars in $p^l_V$ and $\hat{p}^l_V$.
An illustration of a projected 1D Wasserstein distance is shown in figure \ref{fig:sliced-wasserstein}.
\begin{equation}
    \mathcal{L}_{SWD} = \sum_{l=1}^{L} \mathbb{E}_V[ \mathcal{L}_{SW1D}(p^l_V, \hat{p}^l_V) ]
\label{eq:sw}
\end{equation}

\begin{equation}
    \mathcal{L}_{SW1D}(p^l_V, \hat{p}^l_V) = \frac{1}{|p^l_V|} \lVert \texttt{sort}(p^l_V) - \texttt{sort}(\hat{p}^l_V) \rVert^2
\label{eq:sw1d}
\end{equation}

Expectation over random projections $V$ provides a good approximation in practice and an optimized distribution is proven to converge to the target distribution. SWD is known to capture the complete target distribution \cite{pitie2005n} as described below:

\begin{equation}
    \mathcal{L}_{SW}(I, \hat{I}) = 0 \implies p^l = \hat{p}^l , \forall l \in {1, \ldots, L}
\label{eq:swd_prop}
\end{equation}

The calculation of SWD scales in $\mathcal{O}(M \log M)$ for an $M$-dimensional distribution, making it suitable for machine learning applications with gradient descent algorithms.


\subsection{Style Blending.}
Given two different stylized views $I_1, I_2$ and their corresponding feature distributions $\{p_1^l, p_2^l\}$, one may obtain a style-blended scene by refining the source NeRF model towards the Wasserstein barycenter where $t \in [0,1]$ is the blending weight between the two styles:

\begin{equation}
    \mathcal{L}_{style}(I_1, I_2, \hat{I}) =  \sum_{l=1}^{L} \left( t\mathcal{L}_{SWD}(p_1^l, \hat{p}^l) + (1-t) \mathcal{L}_{SWD}(p_2^l, \hat{p}^l) \right)
\label{eq:swd-barycenter}
\end{equation}

\begin{figure*}[t]
  \centering
  \includegraphics[width=\textwidth]{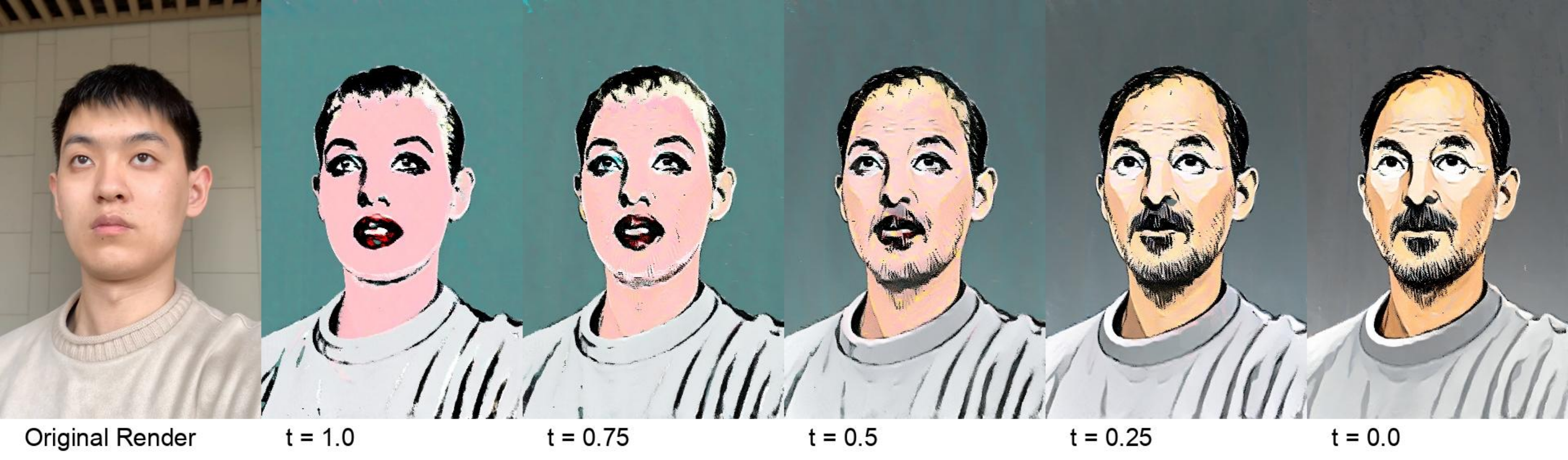}
  \caption{ \textbf{Style Interpolation.} An example of style blending using the Wasserstein barycenter between two different style prompts \textit{"A person like Marilyn Monroe, pop art style"} and \textit{"A person like Steve Jobs"}.
  }
  \label{fig:blend-face}
\end{figure*}

An example of style blending is shown in figure \ref{fig:blend-face}.

\subsection{Implementation Details.}
We employ Stable Diffusion XL \cite{podell2023sdxl} as a backbone for the style-aligned image-to-image diffusion pipeline. For NeRF representation, we use the "nerfacto" model implemented in NeRFStudio \cite{tancik2023nerfstudio}. 
Due to memory constraints, we generate up to $18$ views simultaneously with a fixed seed across all $N$ source views. 
In our experiments, we generated the target images with $50$ denoising steps using a range of classifier-free guidance weights (mostly between 5 and 30) depending on the scene or the style text. While we use depth maps rendered by NeRF for relatively compact and forward-facing scenes, we opt for depth estimations from the MiDaS model \cite{ranftl2020towards} for large-scale outdoor scenes.

As the image editing is performed before NeRF training, our method allows users to test with different text prompts and parameters (\textit{e.g.}, text guidance scale, SDEdit strength) beforehand. Additionally, our straight-forward NeRF training without Iterative DU \cite{instructnerf2023} or score distillation sampling (SDS) loss \cite{poole2022dreamfusion} allows the training process to run with less GPU memory as editing by a diffusion model is not necessary during NeRF updates.
We also verify the importance of style-alignment in the ablation study. Please refer to the supplementary material for more implementation details.
The overall pipeline of our method is shown in Figure \ref{fig:pipeline}.

\begin{figure*}[h]
  \centering
  \includegraphics[width=\textwidth]{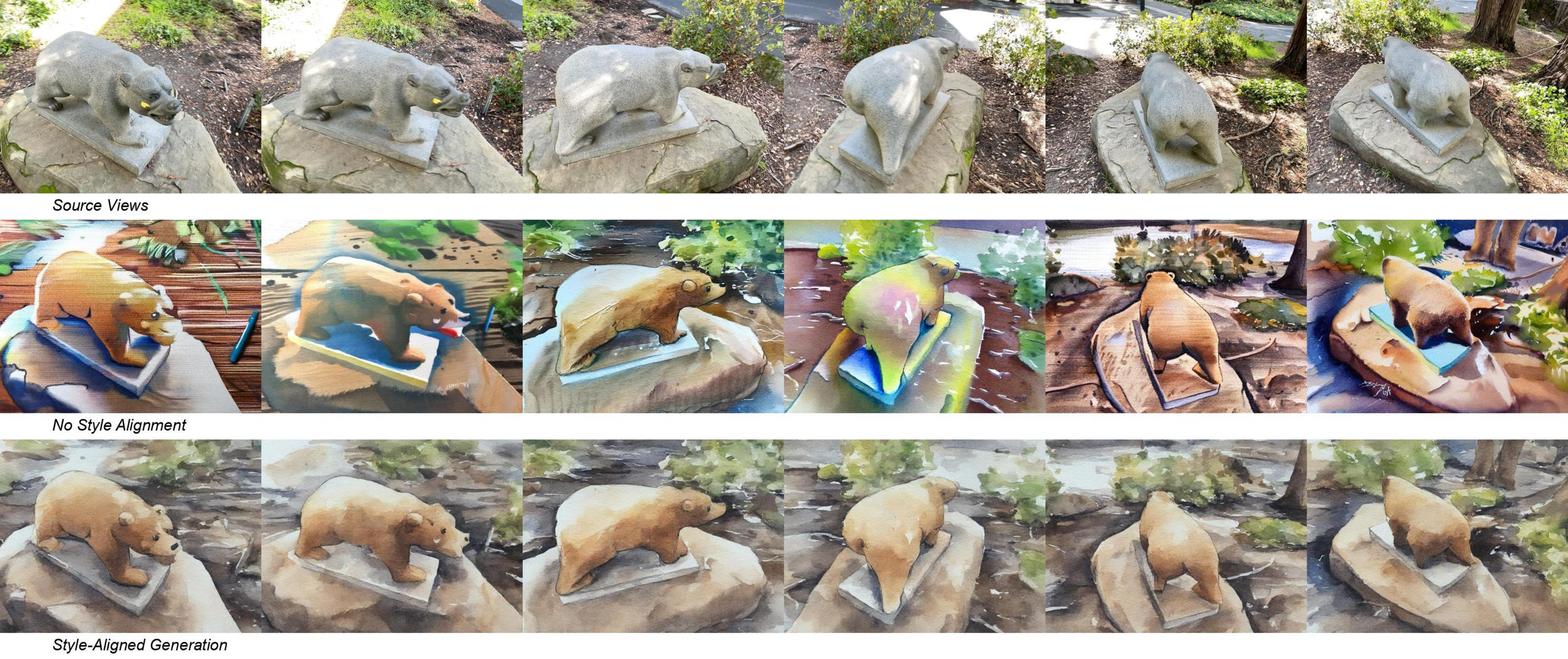}
  \caption{ \textbf{Effect of Style-Alignment.} An example of source view conversion applied to \textit{"Bear"} scene using a text prompt \textit{"A water painting of a brown bear"} with and without shared-attention mechanism within the diffusion pipeline. We find that a fully-shared-attention variant of the style-aligned diffusion model \cite{hertz2023style} greatly improves \emph{style} consistencies among generated views.
  }
  \label{fig:style-align-vs-no-style-align}
\end{figure*}


\section{Results}
We run our experiments on several real-world scenes, including the Instruct-NeRF2NeRF \cite{instructnerf2023} dataset captured by a smartphone and a mirrorless camera with camera poses extracted by COLMAP \cite{schoenberger2016sfm} and PolyCam \cite{polycam}. The dataset contains large-scale 360 scenes, objects, and forward-facing human portraits.

We show qualitative results and comparisons against several variants to verify the effectiveness of our method design with CLIP Text-Image Direction Similarity (CLIP-TIDS), a metric introduced initially in StyleGAN-Nada \cite{gal2022stylegan} and CLIP Directional Consistency (CLIP-DC), a score proposed by Instruct-NeRF2NeRF \cite{instructnerf2023} that aims to measure the directional similarity between original and stylized views. We also evaluate the temporal view consistency \cite{lai2018learning} of the stylized 3D scenes by calculating the average warping error between adjacent frames using FlowNet2 \cite{ilg2017flownet}, an off-the-shelf optical flow estimation model.
In addition to the above, we present comparison results against recent NeRF-based 3D editing methods, Instruct-NeRF2NeRF \cite{instructnerf2023} and ViCA-NeRF \cite{dong2024vica}.
We encourage our readers to see the results in the supplementary video.

\begin{figure}[h]
  \includegraphics[width=\columnwidth]{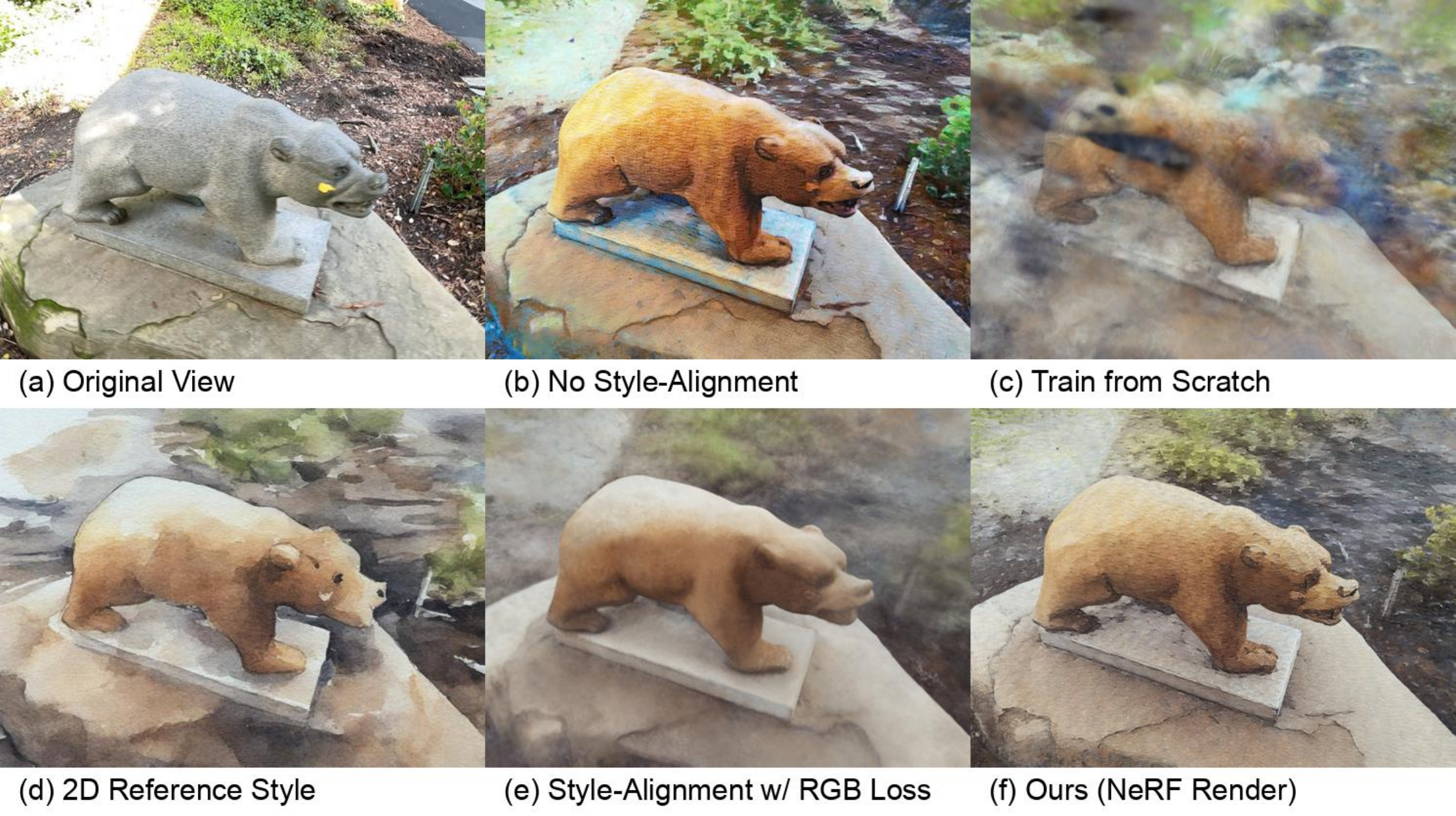} 
  \caption{ \textbf{Baseline Comparisons.} We compare our method against several variants. The images show an example comparison of the "\textit{Bear}" scene trained from a style description "\textit{A water painting of a brown bear}" with a text guidance scale of $7.5$. Note that (b), (c), (e), and (f) are all novel view renders from NeRF. NeRF renderings from (f) ours preserve the original content in (a) without noticeable artifacts compared to (c) Train-from-Scratch and (e) Style-Alignment w/RGB Loss, and also maintain style and color similar to the 2D reference (d). Unlike ours, No Style-Alignment (b) fails to preserve consistent scene color. We encourage our readers to check the results in the video.
  }
  \label{fig:ablation}
\end{figure}

\begin{figure*}[t]
  \centering
  \includegraphics[width=\textwidth]{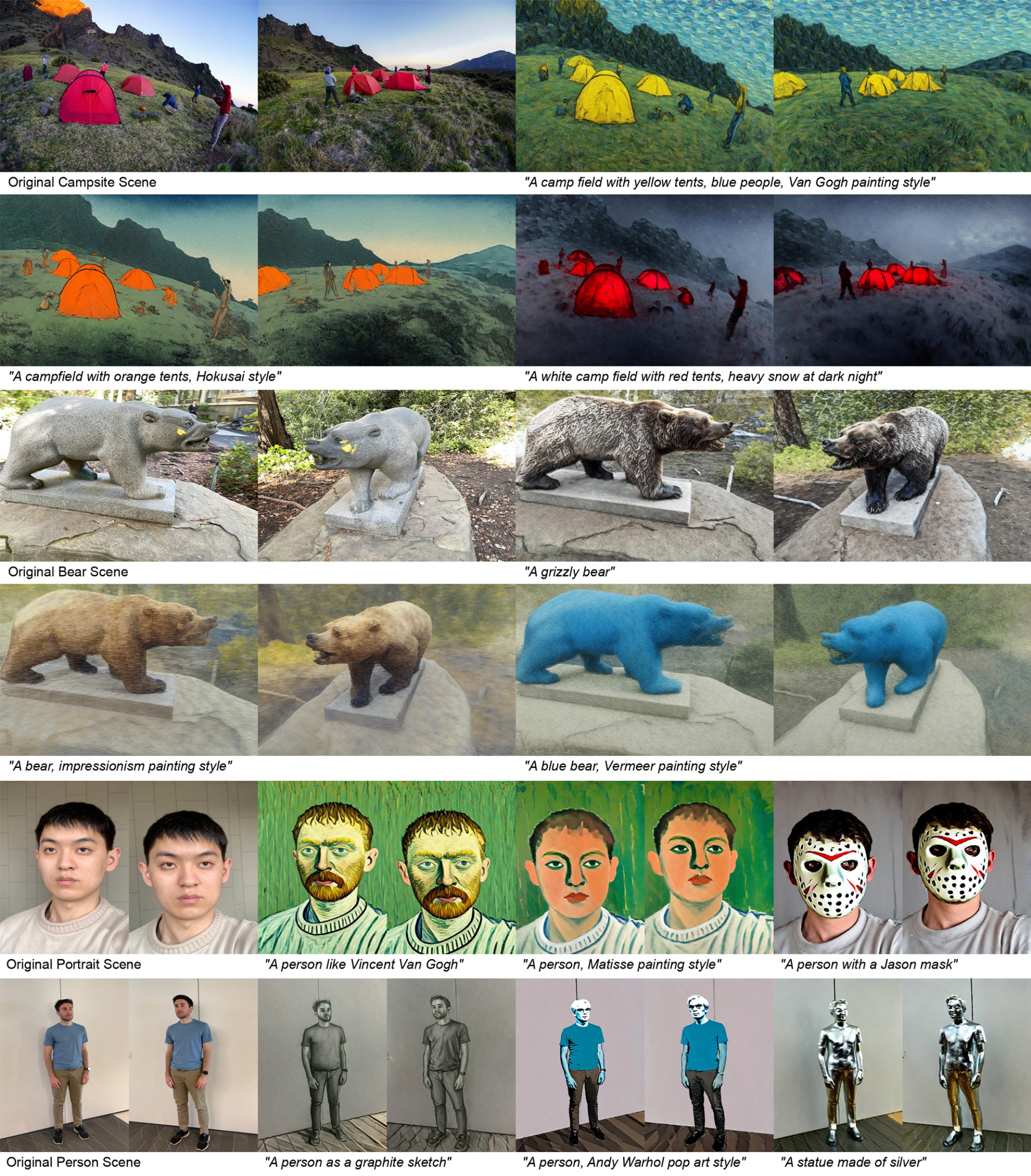}
  \caption{ \textbf{Qualitative Results.} We show novel view rendering examples of real-world scenes stylized or edited with text descriptions specifying certain artistic styles or environmental changes such as weather conditions.
  }
  \label{fig:stylization-examples}
\end{figure*}

\begin{figure*}[t]
  \centering
  \includegraphics[width=\textwidth]{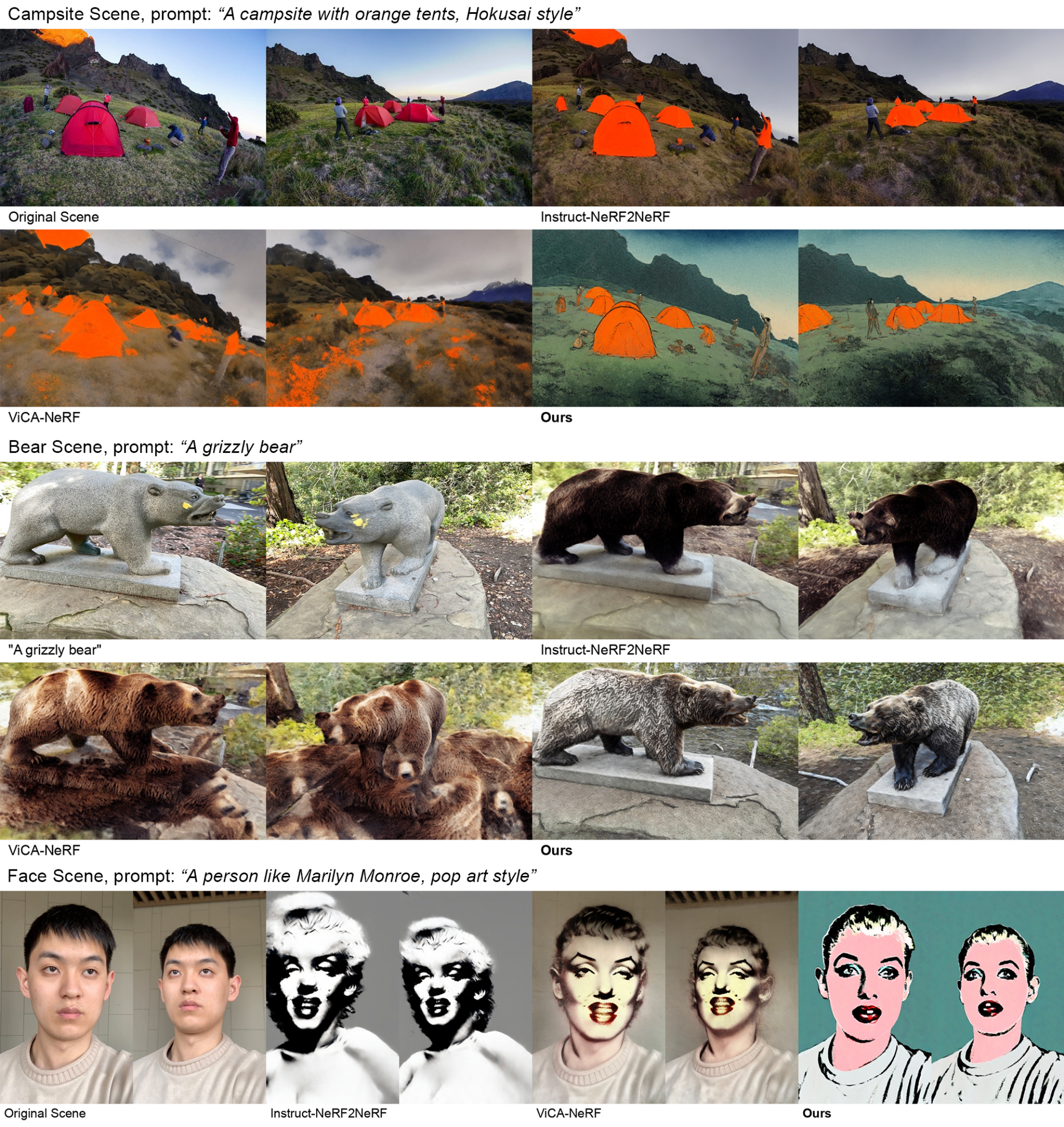}
  \caption{ \textbf{Method Comparison.} A comparison of NeRF stylization methods. While we used a text guidance scale of between $15$ to $25$ for our results, it is controllable via text prompts concerning subjective preferences. Note that all images are novel view renders from NeRF.
  }
  \label{fig:method-comparison}
\end{figure*}

\subsection{Qualitative Evaluation}
Qualitative results are shown in Figure \ref{fig:stylization-examples}. Our method is capable of performing artistic style transfer under various style prompts without hallucinations.
We recommend watching the supplementary video to confirm that the stylized scenes are sufficiently view-consistent.

\subsection{Ablations}
We verify the effectiveness of our method by comparing it against the following variants. An illustration of the comparison results is shown in Figure  \ref{fig:ablation}.
\begin{itemize}
\item \textbf{No Style-Alignment}: To examine the importance of preparing \emph{perceptually} view-consistent stylized images prior to the training process of a source NeRF model, we turn off the full-attention-sharing. Due to the view-inconsistencies in stylized images (See also middle row in figure \ref{fig:style-align-vs-no-style-align}), fine-tuning NeRF on such images will result in an unpredictable mixture of styles.
\item \textbf{Style-Alignment Train-from-Scratch}: In this naive variant, we train a NeRF from scratch using the images generated with our style-aligned diffusion pipeline. Without pre-training of the underlying scene, 3D style transfer produces floating artifacts and shape inconsistencies due to ambiguities in geometry and color of stylized training images.
\item \textbf{Style-Alignment w/ RGB Loss}: This variant trains the source NeRF with $L^1$ pixel RGB loss instead of SWD loss. As \emph{perceputal} view-consistency or similar style does not guarantee physically consistent geometry and color across different views, training with RGB loss tends to diverge to a blurry scene.
RGB loss is prone to over-fitting, whereas SWD is a more valid choice for effectively learning the perceptual similarity from style-aligned training images.
\end{itemize}

\subsubsection{Quantitative Evaluation.} We quantitatively measure our method against the variants using CLIP-TIDS, CLIP-DC, and the average warping error with a fixed text guidance scale of 15. The results are shown in table \ref{tab:clip}. 

\begin{table}[ht]
  \caption{ \textbf{Quantitative Evaluation.} We show CLIP-TIDS, CLIP-DC, and the averaged warping error (MSE) measured across rendered view frames from novel camera trajectories. The values are the average of two scenes using five prompts.
  \label{tab:clip}
  }
  \label{tab:quantitative-evaluations}
  \centering
  \begin{tabular}
{l@{\hspace{2mm}}|@{\hspace{2mm}}c@{\hspace{2mm}}c@{\hspace{2mm}}@{\hspace{2mm}}c@{\hspace{2mm}}c}
    \toprule
     & \vtop{\hbox{\strut No}\hbox{\strut Style-Align}} & \vtop{\hbox{\strut Train from}\hbox{\strut Scratch}} & \vtop{\hbox{\strut Style-Align}\hbox{\strut w/ RGB Loss}} & \textbf{Ours} \\
    \midrule
    CLIP-TIDS $\uparrow$ & $0.125$ & $0.073$ & $0.127$ & $\mathbf{0.162}$ \\
    CLIP-DC $\uparrow$ & $\mathbf{0.932}$ & 0.917 & 0.917 & 0.928 \\
    Warp Error $\downarrow$ & 0.345 & 0.367 & 0.351 & $\mathbf{0.337}$ \\
  \bottomrule
  \end{tabular}
\end{table}

\subsection{Method Comparison}
We compare our method against Instruct-NeRF2NeRF \cite{instructnerf2023} and ViCA-NeRF \cite{dong2024vica} on four scenes including two large-scale outdoor scenes, a 360 object scene (Bear) and a forward-facing scene (a human portrait) using three different text prompts for each scene. Our method exhibits competitive style transfer results whereas previous methods occasionally suffer from hallucination effects (\textit{e.g. The Janus problem} etc...) caused by the underlying diffusion model. As the generation of images and NeRF refinement is a separate process in our method, it is possible to filter out and recreate any images that could have undesired impact on the NeRF fine-tuning.
Visual results are given in figure \ref{fig:method-comparison}.

\subsubsection{User Study}
For each scene, participants were shown a combination of stylized views rendered by different methods in random order, and were asked to select a single view that most likely adhere to the provided stylization text prompt. In our user study, we collected feedbacks from 33 individuals resulting in a total of 396 votes. The overall percentage of the selected preferred method is shown in table \ref{tab:method-comparison}, indicating that our method can perform competitive artistic style transfer without hallucinations.

\subsubsection{Quantitative Comparison}
As style transfer is inherently a subjective task, we think that qualitative evaluation by the user is the most important. Nevertheless, we additionally provide quantitative comparison results using CLIP-TIDS and CLIP-DC. Results are included in table \ref{tab:method-comparison}.

\begin{table}[ht]
  \caption{ \textbf{Method Comparison Results.} The metrics are the average of novel view renders over four scenes with each using three prompts. (Average of $4 \times 3 = 12$ style transfer results.) Our method shows the best values for CLIP-TIDS, CLIP-DC, and user preference.
  \label{tab:method-comparison}
  }
  \centering
  \begin{tabular}
{l|c c c}
    \toprule
     & CLIP-TIDS $\uparrow$ & CLIP-DC $\uparrow$ & User Pref. \\
    \midrule
    Instruct-NeRF2NeRF & $0.081$ & $0.871$ & $15.3\%$ \\
    ViCA-NeRF & $0.061$ & $0.914$ & $17.6\%$ \\
    \textbf{Ours} & \textbf{0.084} & \textbf{0.923} & $\mathbf{67.1\%}$ \\
  \bottomrule
  \end{tabular}
\end{table}

\section{Limitations and Future Work}
While our method may apply artistic style transfer to various 3D scenes, including large-scale outdoor environments, there are several limitations to be considered.
Depending on the strength of the stylization, there may be minor differences between stylized training images and NeRF renderings due to texture variations in the stylized multi-view images (See figure \ref{fig:ablation} (d) vs (f)). While a guidance scale of around 7.5-22.5 produces plausible results, a trade-off exists between stylization strength and view consistency.
Thin structures such as plants and trees in the background or delicate texture patterns are also challenging to reconstruct due to ambiguities in the stylized multi-view images. For the same reason, our method will struggle to learn fine details if there is too much variation in the training images (\textit{e.g.} different people or objects in the background, random patterns of clouds in the sky).
As the style-aligned diffusion pipeline is conditioned on depth maps, significant editing of geometry is also difficult.

We think our approach is applicable to other types of 3D representations such as 3D Gaussian Splatting \cite{10.1145/3592433} and extendable to more features such as scene relighting and deformation, which are exciting directions for further exploration.

\section{Conclusion}
We propose a novel 3D style-transfer method for NeRF representation leveraging a style-aligned generative diffusion pipeline.
By guiding the training process with Sliced Wasserstein Distance or SWD loss, the source 3D scene, pre-trained as a NeRF model, is effectively translated into a stylized 3D scene. 
The method is a relatively straightforward two-step process, allowing the creators to visually search and refine their style concepts by testing various text prompts and guidance scales before fine-tuning the source NeRF model.
Our proposed method shows competitive 3D style transfer results compared to previous methods and can blend styles by optimizing the source 3D scene towards the Wasserstein barycenter.

\begin{acks}
This work was partially supported by JST Moonshot R\&D Grant Number JPMJPS2011, CREST Grant Number JPMJCR2015, and Basic Research Grant (Super AI) of Institute for AI and Beyond of the University of Tokyo. 
We want to thank Instruct-NeRF2NeRF \cite{instructnerf2023} authors for sharing their dataset, and Yuki Kato, Shinji Terakawa, and Yoshiaki Tahara for assisting with data capturing.
\end{acks}


\bibliographystyle{ACM-Reference-Format}
\bibliography{bibliography}

\clearpage

\appendix

\section{Additional Implementation Details}
\label{sec:additional-implementation}
\subsection{NeRF Pipeline}
We pre-train the "nerfacto" model implemented in Nerfstudio \cite{tancik2023nerfstudio} for 60,000 iterations and then fine-tune for 15,000 iterations. We use the default "nerfacto" losses; RGB pixel loss $\mathcal{L}_{rgb}$, distortion loss $\mathcal{L}_{dist}$ \cite{barron2022mip}, interlevel loss $\mathcal{L}_{inter}$ \cite{barron2022mip}, orientation loss 
$\mathcal{L}_{orien}$ \cite{verbin2022ref}, and predicted normal loss $\mathcal{L}_{normal}$ \cite{verbin2022ref} for pre-training (equation \ref{eq:pre-train}). During fine-tuning, we disable the RGB pixel loss $\mathcal{L}_{rgb}$, orientation loss $\mathcal{L}_{orien}$, and predicted normal loss $\mathcal{L}_{normal}$ but add the Sliced Wasserstein Distance (SWD) loss \cite{heitz2021sliced} (equation \ref{eq:fine-tune}). The total loss function for each phase is as follows:

\begin{equation}
    \mathcal{L}_{pre} = \mathcal{L}_{rgb} +  \lambda_1 \mathcal{L}_{dist} + \lambda_2 \mathcal{L}_{inter} + \lambda_3 \mathcal{L}_{orien} + \lambda_4 \mathcal{L}_{normal}
\label{eq:pre-train}
\end{equation}

\begin{equation}
    \mathcal{L}_{fine} = \mathcal{L}_{swd} +  \lambda_1 \mathcal{L}_{dist} + \lambda_2 \mathcal{L}_{inter}
\label{eq:fine-tune}
\end{equation}
where we use the default hyper-parameters $\lambda_1 = 0.002, \lambda_2=1.0, \lambda_3 = 0.0001, \lambda_4=0.001$ for most cases. A greater value for the distortion loss may work better if floating artifacts remain in the scene. We list a brief description of the "nerfacto" losses.

\begin{itemize}
    \item \textbf{Distortion Loss:} The loss encourages the density along a ray to become compact, aiming to prevent floaters and background collapse. It was proposed in Mip-NeRF 360 \cite{barron2022mip}.
    \item \textbf{Interlevel Loss:} The loss allows the histograms of the point sampling proposal network and NeRF network to become more consistent. It was also proposed in Mip-NeRF 360 \cite{barron2022mip}
    \item \textbf{Orientation Loss:} The loss aims to prevent "foggy" surfaces by penalizing visible samples with predicted normals facing the ray direction. It was introduced in Ref-NeRF \cite{verbin2022ref}.
    \item \textbf{Predicted Normal Loss:} The loss enforces the predicted normals to be consistent with density gradient normals. It is often used in conjunction with the orientation loss.
\end{itemize}
For detailed definitions of the "nerfacto" losses, please see Mip-NeRF \cite{barron2022mip} and Ref-NeRF \cite{verbin2022ref}. 

We run all our experiments with Python 3.10 and CUDA 11.8 on a single NVIDIA H100. Large-scale outdoor 360 scenes (such as "Campsite" or "Farm," consisting of 174 multi-view training images) takes -15 minutes for pre-training with -6GB GPU memory and fine-tuning takes -30 minutes with -20GB GPU memory (which depends on the number of ray samples).

The SWD loss $\mathcal{L}_{swd}$ is applied to $64 \times 64$ patches sampled during NeRF fine-tuning. Although we found that $64 \times 64$ empirically works sufficiently, one may change the patch size accordingly.
While we optionally enable SDEdit \cite{meng2021sdedit} in our style-aligned diffusion pipeline, we recognized that depth maps are often enough for conditioning on the original views. In such cases, we use $1.0$ as the strength for SDEdit.

\subsection{SWD Implementation}
We follow the implementation of \cite{heitz2021sliced} using the first 12 layers of VGG19 with uniformly sampled random projections for the SWD calculation.

\section{Additional Comparison of Loss Functions}
\begin{figure*}[t]
  \centering
  \includegraphics[width=\textwidth]{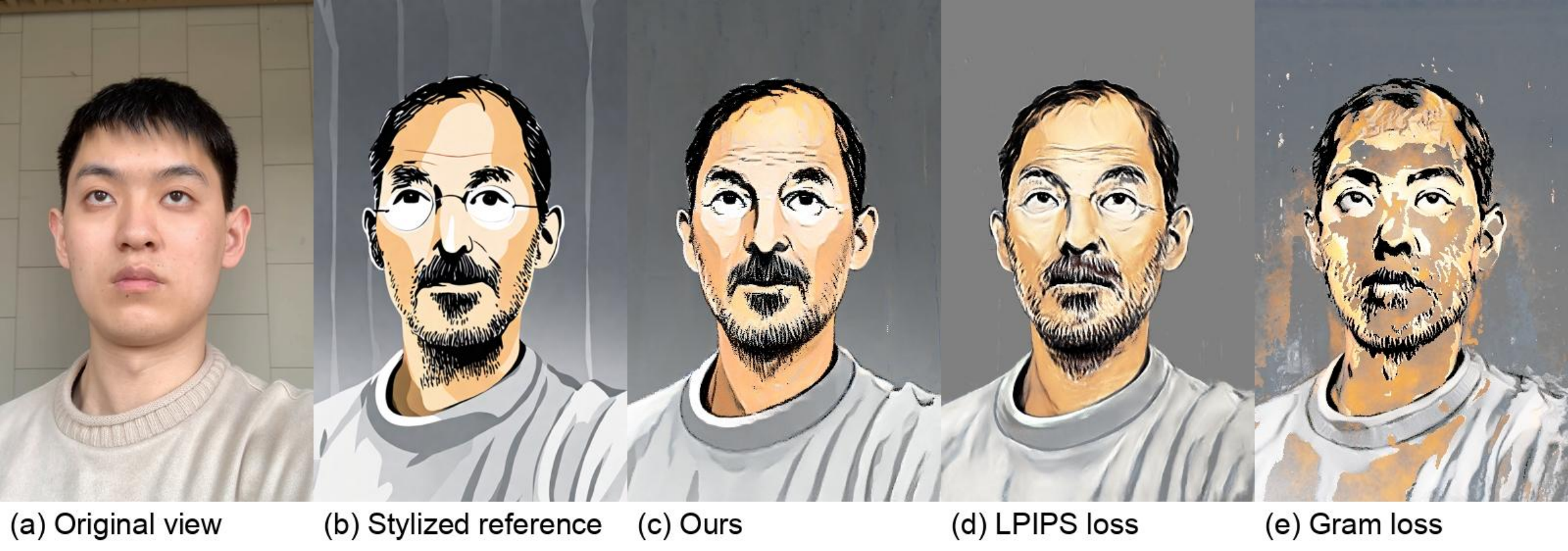}
  \caption{ \textbf{Loss function comparison.} We show novel view renders from fine-tuned NeRF models trained with different loss functions: (c) Ours (SWD Loss), (d) LPIPS Loss, and (e) Gram Loss. Gram loss introduces noticeable artifacts. While the LPIPS variant performs better than the Gram loss version, our NeRF render results are more similar to the stylized reference image with fewer artifacts.
  }
  \label{fig:comparison-loss}
\end{figure*}

We provide comparisons against two related loss functions: the Gram loss and Learned Perceptual Image Patch Similarity (LPIPS), and discuss the relative effectiveness of the SWD loss. Please see figure \ref{fig:comparison-loss} for a visual comparison. We also show quantitative evaluation results in table \ref{tab:quantitative-loss-comparison}.

Since style is known to be well represented by the feature maps of VGG19, we are interested in a loss term that accurately captures their distributions. Given the sets of feature distributions $p$ and $\hat{p}$ for the corresponding images $I$ and $\hat{I}$, the formal objective is to employ a loss term $\mathcal{L}$ satisfying equation \ref{eq:feature_prop} where $l$ denotes the layer number of VGG19.
In short, we choose SWD loss (\textit{i.e.}, $\mathcal{L} = \mathcal{L}_{SW}$) as it can capture the complete stationary statistics of VGG19 feature distributions. 

\begin{equation}
    \mathcal{L}(I, \hat{I}) = 0 \implies p^l = \hat{p}^l , \forall l \in {1, \ldots, L}
\label{eq:feature_prop}
\end{equation}

\subsection{Gram Loss}
Gram loss introduced by \cite{gatys2015neural} is defined as the mean-squared error between Gram matrices of the feature distributions:

\begin{equation}
    \mathcal{L}_{Gram}(I, \hat{I}) = \sum^{L}_{l}\frac{1}{N^2_{l}} \left\| G^l - \hat{G}^l \right\|^2
\label{eq:gram_loss}
\end{equation}
where $G^l$ and $\hat{G}^l$ denote the Gram matrices of feature maps from images $I$ and $\hat{I}$ at layer $l$.
Given a feature map of $M_l$ pixels with $N_l$ channels, an element $G_{i,j}$ of the Gram matrix $G^l \in \mathbb{R}^{N_l \times N_l}$ is defined as the second-order cross moment between features at channel $i$ and $j$.

\begin{equation}
    G^l_{i,j} = \frac{1}{M_l} \sum^{M_l}_{m=1} F^l_m[i]F^l_m[j]
\label{eq:gram_element}
\end{equation}

Although Gram loss is often utilized as a convenient style loss for its capability to capture the feature statistics, Gram loss cannot capture the full distribution of features, resulting in some artifacts, whereas Wasserstein loss is able to capture the complete target distribution.

\begin{equation}
    \mathcal{L}_{Gram}(I, \hat{I}) = 0 \rlap{\(\quad\not\)}\implies p^l = \hat{p}^l , \forall l \in {1, \ldots, L}
\label{eq:gram_prop}
\end{equation}

\subsection{LPIPS}
LPIPS \cite{zhang2018unreasonable} is a metric developed for measuring perceptual similarity that well agrees with human evaluations.
LPIPS is calculated with a pre-trained model that takes averaged feature maps as input and is trained via cross-entropy loss based on human-judged data. While LPIPS excels at capturing the perturbation-
invariant perceptual similarity of patches, SWD better represents
the raw VGG19 feature distributions, which makes it more appropriate for style transfer tasks. Replacing SWD with LPIPS produces mild artifacts.

\begin{table}[ht]
  \caption{ \textbf{Additional Quantitative Comparison.} We show CLIP-TIDS, CLIP-DC, and the averaged warping error (MSE) measured across rendered view frames from novel camera trajectories. The values are the average of two scenes using five prompts.
  }
  \label{tab:quantitative-loss-comparison}
  \centering
  \begin{tabular}
{l@{\hspace{2mm}}|@{\hspace{2mm}}c@{\hspace{2mm}}@{\hspace{2mm}}c@{\hspace{2mm}}c}
    \toprule
     & Gram Loss & LPIPS Loss & \textbf{Ours} (SWD Loss) \\
    \midrule
    CLIP-TIDS $\uparrow$ & $0.127$ & $0.115$ & $\mathbf{0.162}$ \\
    CLIP-DC $\uparrow$ & 0.868 & 0.887 & $\mathbf{0.928}$ \\
    Warp Error $\downarrow$ & 0.323 & 0.332 & $\mathbf{0.337}$ \\
  \bottomrule
  \end{tabular}
\end{table}

\section{CLIP Text-Image Direction Similarity}
CLIP Text-Image Direction Similarity (CLIP TIDS) \cite{gal2022stylegan} is a metric for evaluating how well the change in the stylized image is aligned with the user-provided text prompt. Given the CLIP image and text encoder $E_I, E_T$, CLIP TIDS between the source and the stylized image $I_{source}, I_{style}$ is calculated as:

\begin{equation}
    \Delta I = E_I(I_{style}) - E_I(I_{source}), \Delta T = E_T(t_{style}) - E_T(t_{source})
\label{eq:clip-tids}
\end{equation}
\begin{equation}
    \texttt{CLIP-TIDS} \equiv \frac{ \Delta I \cdot \Delta T }{ |\Delta I||\Delta T| }
\label{eq:clip-tids2}
\end{equation}
where $t_{source}, t_{style}$ are text descriptions describing the images (\textit{e.g.} $t_{source}=$"A photo of a person", $t_{style}=$"A person like Vincent Van Gogh").

\section{Effect of Text Guidance Scale}
\begin{figure*}[t]
  \centering
  \includegraphics[width=\textwidth]{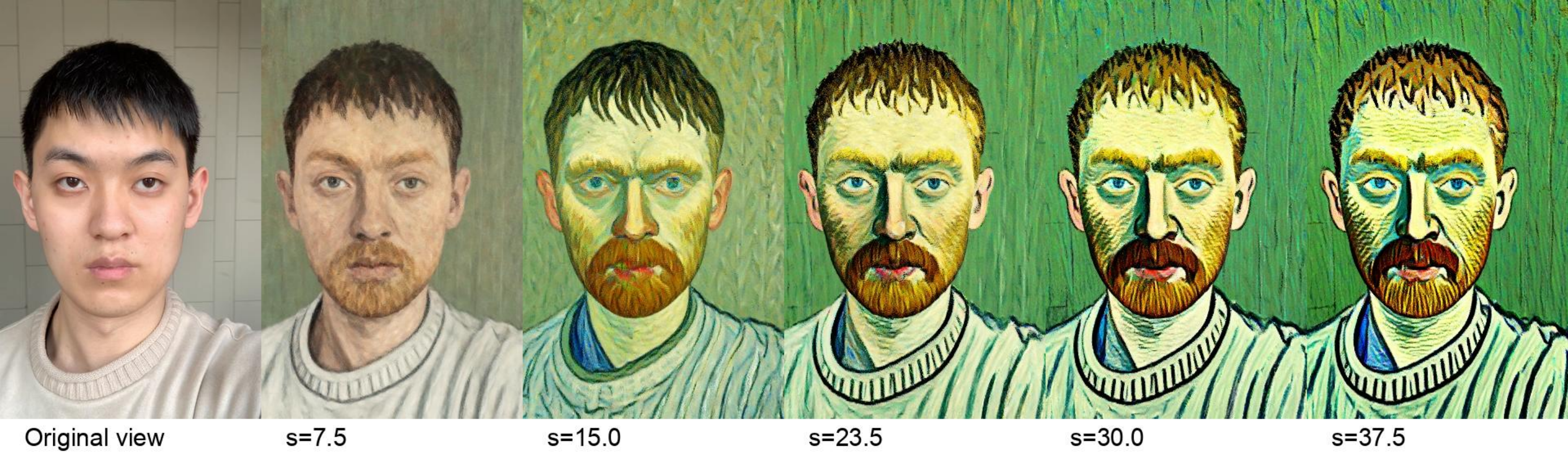}
  \caption{ \textbf{Effect of Text Guidance Scale.} We show some NeRF rendering results for the prompt \textit{"A person like Vincent Van Gogh"} using various text guidance scales ($s=7.5, 15.0, 22.5, 30.0, 37.5$).
  }
  \label{fig:effect-of-scale}
\end{figure*}

In figure \ref{fig:effect-of-scale}, we show an example of NeRF renderings using the same style prompt \textit{"A person like Vincent Van Gogh"} but different text guidance scales. We may see that style strength is controllable while keeping the original content structure grounded on the original image. We also show CLIP TIDS for each text guidance scale in table \ref{tab:scale-clip}. A stronger text guidance scale will lead to higher CLIP TIDS.

\begin{table}[ht]
  \caption{ \textbf{CLIP-TIDS Comparison.} CLIP-TIDS values over test renders for each text guidance scale are shown. We can verify that a stronger text guidance scale results in higher CLIP TIDS values.
  \label{tab:scale-clip}
  }
  \centering
  \begin{tabular}
{l@{\hspace{2mm}}|@{\hspace{2mm}}c@{\hspace{2mm}}c@{\hspace{2mm}}@{\hspace{2mm}}c@{\hspace{2mm}}c@{\hspace{2mm}}c}
    \toprule
     Text Guidance & $s=7.5$ & $s=15.0$ & $s=22.5$ & $s=30.0$ & $s=37.5$ \\
    \midrule
    CLIP TIDS & $0.0693$ & $0.1459$ & $0.1520$ & $0.1594$ & 0.1718\\
  \bottomrule
  \end{tabular}
\end{table}

\begin{figure*}[t]
  \centering
  \includegraphics[width=\textwidth]{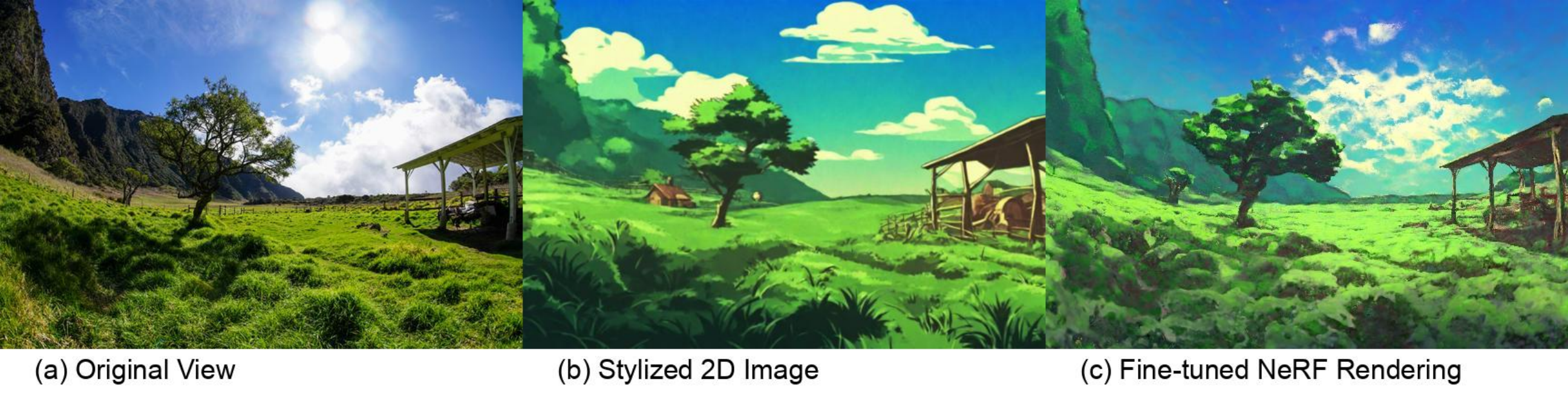}
  \caption{ \textbf{Limitations.} Due to remaining ambiguities in the stylized multi-view images, fluctuating objects such as clouds may lose their detailed shape in the fine-tuned NeRF renderings. We wish to improve on this in our future work.
  }
  \label{fig:limitations}
\end{figure*}

\section{Limitations}
While our method can effectively perform overall style transfer to 3D scenes, it is still difficult to reconstruct a detailed structure of fluctuating objects within the stylized multi-view images. In figure \ref{fig:limitations}, for example, (c) the clouds in the fine-tuned NeRF scene show a fractal-like pattern, which is different from (b) the clouds illustrated in the stylized image. This phenomenon is due to the ambiguities of cloud positions or shapes appearing in the stylized images generated by the style-aligned diffusion pipeline \cite{hertz2023style}. We leave the development of a more robust content structure-preserving style transfer technique as future work.

\end{document}